\documentclass[letterpaper]{article} % DO NOT CHANGE THIS
\usepackage{aaai23}  % DO NOT CHANGE THIS
\usepackage{dsfont}
\usepackage{times}  % DO NOT CHANGE THIS
\usepackage{helvet}  % DO NOT CHANGE THIS
\usepackage{courier}  % DO NOT CHANGE THIS
\usepackage[hyphens]{url}  % DO NOT CHANGE THIS
\usepackage{graphicx} % DO NOT CHANGE THIS
\usepackage{natbib}  % DO NOT CHANGE THIS AND DO NOT ADD ANY OPTIONS TO IT
\usepackage{caption} % DO NOT CHANGE THIS AND DO NOT ADD ANY OPTIONS TO IT
\usepackage{algorithm}
\usepackage{algorithmic}
\usepackage{newfloat}
\usepackage{listings}
\DeclareCaptionStyle{ruled}{labelfont=normalfont,labelsep=colon,strut=off} % DO NOT CHANGE THIS
\floatstyle{ruled}
\newfloat{listing}{tb}{lst}{}
\floatname{listing}{Listing}
\title{In-game Toxic Detection: Bi-directional Representations with Attention Residuals}
\author{
    Yuanzhe Jia
}
\affiliations{
    University of Sydney, Australia\\
    yjia5612@uni.sydney.edu.au
}

\usepackage{bibentry}
\begin{document}

\maketitle

\begin{abstract}
In-game toxic language has emerged as a critical concern in the gaming industry and community. While several frameworks and models for online game toxicity analysis have been proposed, detecting toxicity in player chat utterances remains a formidable challenge: stemming not only from the extremely short length of such utterances but also from the heavy reliance on game slang, abbreviations, and domain-specific jargon, which generic language models are poorly suited to recognize. This paper presents a shared task for in-game toxic language detection built upon real-world in-game chat data, and proposes the best-preforming model for the toxic language slot filling: Bi-directional Representations with Attention Residuals (BRAR). Experimental results demonstrate that BRAR effectively captures the global context and outperforms the existing baselines on slot filling. The relevant code is publicly available on GitHub\footnote{\url{https://github.com/yuanzhe-jia/toxic-language-detection}}.

%In-game toxic language becomes the hot potato in the gaming industry and community. There have been several online game toxicity analysis frameworks and models proposed. However, it is still challenging to detect toxicity due to the nature of in-game chat, which has an extremely short length. In this paper, we describe how the in-game toxic language shared task has been established using real-world in-game chat data. In addition, we propose and introduce the model/framework for toxic language token tagging (slot filling) from the in-game chat. The relevant code is publicly available on GitHub\footnote{\url{https://github.com/yuanzhe-jia/toxic-language-detection}}.
\end{abstract}

\section{Introduction}
Toxic behavior has become a critical concern in online gaming, yet detection is complicated by the distinctive nature of in-game chat: unlike social media or news, it is substantially shorter because players type while playing, with longer utterances confined to pre- or post-game discussions. This brevity, compounded by pervasive game slang, abbreviations, and domain-specific jargon that generic language models often fail to capture, renders slot-level analysis indispensable for detecting in-game toxicity. Slot filling, the core technique underlying such token-level detection, is typically cast as sequence labeling and has progressed from feature-engineered statistical models such as CRF to deep learning approaches, with Mesnil first demonstrating that the bi-directional RNN with CRF decoders can improve the performance~\cite{1}. However, RNNs suffer from vanishing/exploding gradients and fail to capture long-term dependencies explicitly; LSTMs and GRUs mitigate this via gating but still lack explicit dependency modeling and parallel processing. The attention mechanism was therefore adopted, and several sequence-to-sequence models with the attention layer have been applied to slot filling, including encoder–decoder enhancement~\cite{2}, position-aware attention~\cite{3}, joint delexicalized generation and label prediction~\cite{4}, and pointer-network-based slot value prediction~\cite{5}. Nevertheless, few approaches integrate BiLSTM, Attention, and CRF to capture adjacent-word interactions, highlight key words, and exploit label dependencies. To address the gap, this paper describes the established slot (token)-based in-game toxic language detection shared task and presents the best-performing model with its novel components.

%Toxic behavior has become a severe problem in recent online games and the gaming industry. Compared to other online domains, such as social media or online news, it has a much shorter length because game players tend to type in-game chat during playing. The longer utterances occur only in pre- or post-game discussions. With this in-game chat nature in mind, understanding the slot-level~\cite{8} is crucial to detect the in-game toxic language. In this paper, we describe the established slot (token)-based in-game toxic language detection shared task, and present the best model and its novel components. The result shows the novelty of the best model by comparing the baselines from the CONDA~\cite{1}.

\section{Shared Task and Dataset}
We established a shared task competition\footnote{\url{https://www.kaggle.com/competitions/2022-comp5046-a2}} for sequence token labeling, aiming to identify the semantic category of each slot in chat utterances from CONDA~\cite{6}, a dataset comprising 44,869 utterances derived from chat logs of 1,921 Dota 2 matches and annotated with six distinct slot labels—\textbf{T} (Toxicity), \textbf{C} (Character), \textbf{D} (Dota-specific), \textbf{S} (Game Slang), \textbf{P} (Pronoun) and \textbf{O} (Other)—to facilitate a deeper understanding of game context. Given the informal and noisy nature of in-game chat, the task demanded models capable of handling lexical variants and domain-specific terminology. Hosted on Kaggle, the competition attracted 312 teams, each permitted 20 submissions per day, culminating in 3,646 submissions over four weeks, with evaluation based on overall micro-F1 excluding the O tag. To ensure reproducibility, a standard data split and evaluation script were released, and participants were encouraged to report both overall and per-label F1 scores. Most participants preprocessed the provided data by tokenizing utterances into slots using spaces, as CONDA provides cleaned utterances with slot labels, while some employed data augmentation to enlarge their training sets. For input embeddings, participants leveraged combinations of syntactic, semantic, and domain-related representations: syntactic embeddings typically encoded POS tags and/or dependency parsing results from SpaCy, whereas semantic embeddings either adopted pre-trained GloVe/FastText vectors or were trained from scratch using FastText/Word2Vec on the provided game chat corpus. Diverse sequence labeling architectures were explored, and we subsequently present the methodology and experimental results of the best-performing model.

%We set up a shared task competition\footnote{\url{https://www.kaggle.com/competitions/2022-comp5046-a2}} for performing a sequence token labeling task to identify the type of semantics conveyed by each slot in-game chat utterances provided by CONDA~\cite{6}. CONDA consists of 44,869 utterances from chat logs of 1,921 Dota2 matches. CONDA provides 6 distinct slot labels: \textbf{T} (Toxicity), \textbf{C} (Character), \textbf{D} (Dota-specific), \textbf{S} (Game Slang), \textbf{P} (Pronoun) and \textbf{O} (Other) to provide a deeper understanding of game context. A total of 312 teams participated in our shared task. Each team was allowed 20 submissions per day. In total, we received 3,646 submissions across 4 weeks. Most participants pre-processed the provided data simply by tokenizing the utterances into slots by spaces, as CONDA provides the cleaned utterances to provide slot labels. There are also teams applying data augmentation to increase the number of instances they can have for training. For the input embedding, combinations of syntactic, semantics and domain-related embeddings were included in the participants' trials. For the syntactic embedding, most teams chose to encode POS tags and/or dependency parsing results based on models provided by SpaCy. For the semantic embedding, participants would either directly use pre-trained GloVE/FastText embedding vectors, or train their own FastText/Word2Vec based on the provided game chat corpus. A variety of sequence labeling model architectures are explored by the participants, and we are going to introduce the best-performing team's methodology and experimental results.

\section{Methodology}
\begin{figure}[t]
\centering
\includegraphics[width=1.0\columnwidth]{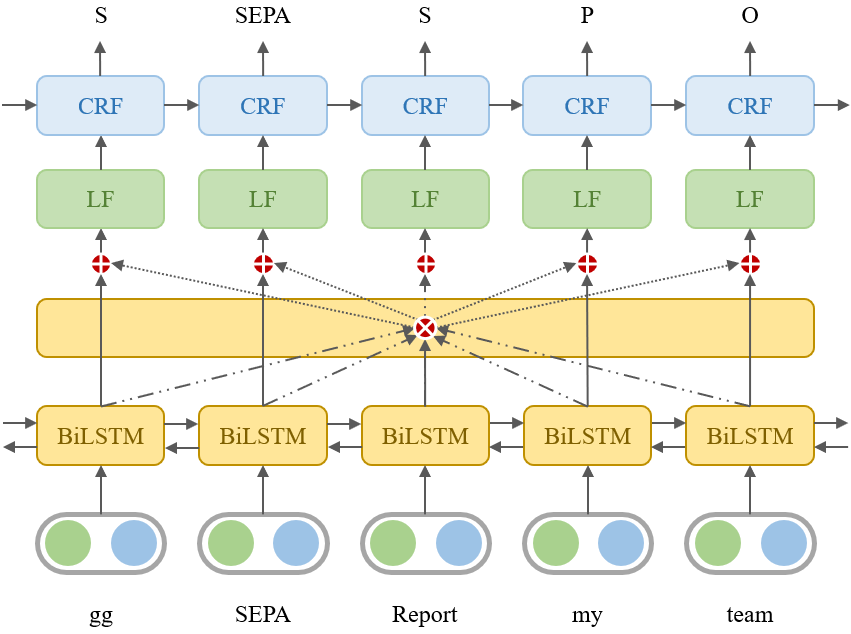}
\caption{The figure illustrates the overall architecture of BRAR, in which the dotted arrows denote the attention residuals and the green boxes (labeled “LF”) present the label forcing technique.}
\label{fig:structure}
\end{figure}

The proposed model is a combination of BiLSTM cells, attention residuals, the label forcing technique, and CRF decoders. Since the model uses the global information extracted from the attention mechanism as residuals to supplement bi-directional features, it is named Bi-directional Representations with Attention Residuals (BRAR). The overall architecture is shown in Figure~\ref{fig:structure}. The input example, “gg SEPA Report my team”, is divided into the sequence $ x=(x_1,x_2,\ldots,x_t) $ where $ t $ denotes the number of tokens. The output of the model is the corresponding slot labels $ y=(y_1,y_2,\ldots,y_t) $ with $ c $ unique values.

\subsection{BiLSTM}
The BiLSTM layer performs feature extraction on the input data in sequence, and the hidden state obtained at each time step is then passed to the attention layer. Assume the hidden size of LSTM is $ \lambda $, the hidden state 
$ h_i \in \mathds{R}^{\lambda} $ 
at time step $ i $ is represented as the abbreviation below:
\begin{equation}\label{eq1}
h_i = LSTM(x_i, h_{i-1})
\end{equation}

And the bi-directional hidden state 
$ h_i \in \mathds{R}^{2\lambda} $ 
that involves the forward hidden state $ h_i^{forward} $ and the backward hidden state $ h_i^{backward} $ is depicted:
\begin{equation}\label{eq2}
h_i = [h_i^{forward}, h_i^{backward}]
\end{equation}

\subsection{Attention Residual}
The attention layer aims to understand the global information and find the main ideas of the input utterance. The attention 
$ a \in \mathds{R}^{2\lambda} $ 
is calculated in the following equation where
$ h_t \in \mathds{R}^{2\lambda} $ 
is the last hidden state of the BiLSTM layer,
$ W_a \in \mathds{R}^{{2\lambda} \times {t}} $ 
is a trainable weight matrix and 
$ H \in \mathds{R}^{{t} \times {2\lambda}} $ 
is output the hidden state of BiLSTM:
\begin{equation}\label{eq3}
a = softmax(h_t \cdot W_a \cdot H) 
\end{equation}

As the global information is not always helpful for understating each input token, it is treated as the residual of the token-level representation to form the feature 
$ f_i \in \mathds{R}^{{t} \times {c}} $,
while $ \alpha $ is a trainable parameter for scaling and
$ W_f \in \mathds{R}^{{2\lambda} \times {c}} $ 
is a trainable weight matrix for dimension transformation. When the global information is beneficial to the token-level interpretation, it will speed up the convergence; otherwise, it will not reduce the prediction performance.
\begin{equation}\label{eq4}
f_i = W_f(H + \alpha \times a_i)
\end{equation}

\subsection{Label Forcing}
The feature representation is enhanced at the label forcing layer to form the emission scores, which will be passed to the CRF layer to predict the tag of each token. As both training and test sets contain a large number of identical words, the classifications of these words are highly likely to be the same in a specific domain. For example, “gg” stands for “good game”. If all the words “gg” in the training set are classified as “S” (Game Slang), then the word in the test set will have a high probability of being classified as “S”. That is to say, if the probability information of the correspondence between words and labels in the training set can be learned, it will be of great help in predicting the results of the test set. Therefore, the proposed model uses the label forcing technique: the label distribution probability of each token in the training corpus is calculated and normalized by the following equation, where $ C_{ij} $ is the frequency of that the token $ i $ is annotated as the label $ j $:
\begin{equation}\label{eq5}
p_i = C_{ij} / \sum_j C_{ij}
\end{equation}

\begin{figure}[t]
\centering
\includegraphics[width=1.0\columnwidth]{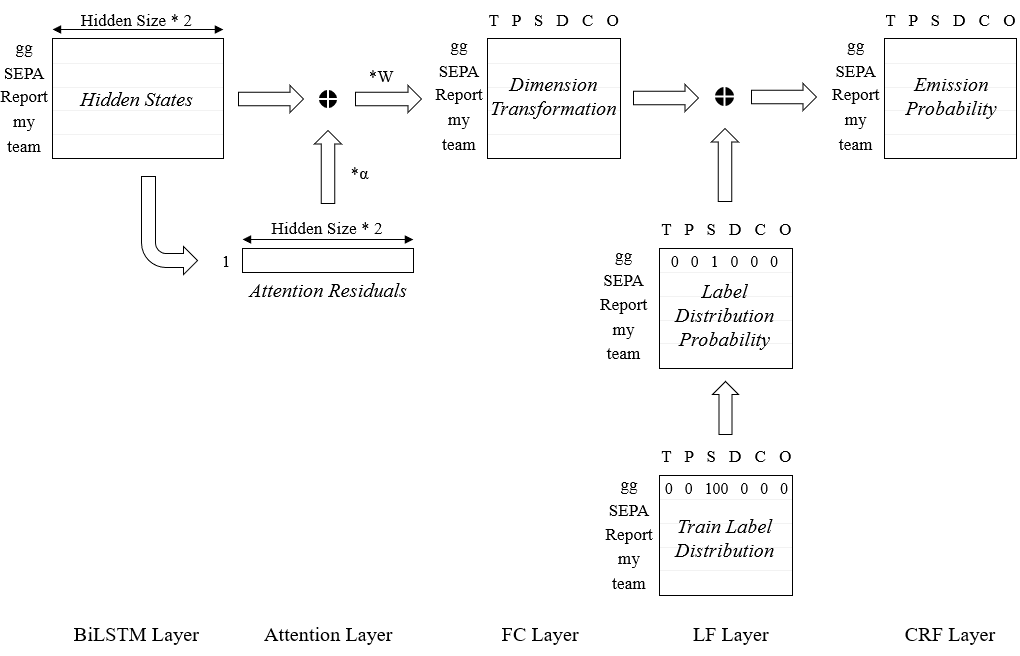}
\caption{The figure illustrates the process by which the parameter matrices in BRAR undergo a series of dimensional transformations and ultimately yield the emission scores for the CRF layer.}
\label{fig:emission}
\end{figure}

After that, the label distribution probability $ p_i $ will be added element-wise along with the feature representation $ f_i $ to form the emission scores
$ e_i \in \mathds{R}^{{t} \times {c}} $ 
for the CRF layer (Ref. Figure~\ref{fig:emission}):
\begin{equation}\label{eq6}
e_i = f_i + p_i
\end{equation}

\subsection{CRF}
Given that slot label predictions are inherently interdependent, modeling the transitional dependencies among neighboring labels is beneficial for producing coherent tag sequences. Accordingly, a CRF layer is placed atop the proposed model to jointly decode the optimal chain of labels for the utterance. This design enables the model to enforce valid tag transitions and prevents locally optimal but globally inconsistent predictions.

\section{Experiment}
The proposed model adopts FastText-50 as input embeddings, with a hidden size of 5 per direction in the BiLSTM layer and the number of layers fixed at 1. The SGD optimizer is employed with a learning rate of 0.1 and a weight decay of 1e-4, and training proceeds for 2 epochs with a batch size of 1, following the CONDA setting. Evaluation metrics include the overall micro-F1 and per-slot F1 on the test set, excluding the O tag when computing the overall F1, and five baselines are taken from the CONDA paper. We first conducted an ablation study on the number of stacked BiLSTM layers. As presented in Table~\ref{tab:ablation}, increasing the number of layers leads to lower F1 scores, although the differences among the three configurations are not statistically significant. This suggests that deeper architectures tend to overfit the relatively short and noisy in-game chat utterances. Consequently, we use a single BiLSTM layer owing to its lower parameter count and computational efficiency. In comparison with the existing baselines, Table~\ref{tab:comparison} shows that BRAR achieves superior F1 scores across most labels, especially for the T, S, and D labels, attributable to the attention residuals and the label forcing technique that enhance the capture of global context and high-frequency label patterns.

%The proposed model adopts FastText-50 as the input embeddings. The BiLSTM layer contains a hidden size of 5 in each direction, and the layer number is set to 1. The SGD optimizer has been applied with the learning rate of 0.1 and the weight decay of 1e-4. In addition, the training epoch is set to 2 with a batch size of 1 (same as the CONDA). As for evaluation metrics, the overall micro F1 and that for each slot label have been reported on the test set, but the O tag is excluded when calculating the overall F1-Score. The 5 baselines are selected based on CONDA paper. From Table \ref{tab:comparison} we can conclude that BRAR outperforms other baselines on most F1, especially in predicting T, S, and D labels. We evaluated the model with different number of Bi-LSTM stacks/layers. The results shown in Table \ref{tab:ablation} indicate that more layers lead to the lower F1 scores, though there is no significant difference among the three different number of layers. Therefore, we propose to use only 1 BiLSTM layer as it is computationally more efficient with a smaller number of parameters.

\begin{table}
\centering
\begin{tabular}{cccc cccc}
\hline\noalign{\smallskip}	
Number of Stacks & F1 & F1(T) & F1(S) & F1(D) \\
\noalign{\smallskip}\hline\noalign{\smallskip}
1 layer & \textbf{99.9} & \textbf{98.6}  & \textbf{99.4} & \textbf{98.1}  \\
2 layers & 99.6 & 98.1 & 99.3 & 96.4 \\
3 layers & 98.6 & 96.5 & 98.3 & 81.8 \\
\noalign{\smallskip}\hline
\end{tabular}
\caption{Ablation study with different stacks (\%).}
\label{tab:ablation} 
\end{table}

\begin{table}
\centering
\begin{tabular}{cccc cccc}
\hline\noalign{\smallskip}	
Model & F1 & F1(T) & F1(S) & F1(D) \\
\noalign{\smallskip}\hline\noalign{\smallskip}
RNN-NLU~\shortcite{7} & 97.0 & 93.1  & 93.0 & 71.8 \\
Slot-gated~\shortcite{8} & 99.1 & 97.8  & 98.2 & 95.2 \\
Inter-BiLSTM~\shortcite{9} & 86.5 & 87.1  & 86.9 & 78.8 \\
Capsule NN~\shortcite{10} & 99.1 & 97.5  & 98.2 & 94.9 \\
Joint BERT (2019) & 98.9 & 97.2  & 97.9 & 91.4 \\
BRAR (our model) & \textbf{99.9} & \textbf{98.6}  & \textbf{99.4} & \textbf{98.1} \\
\noalign{\smallskip}\hline
\end{tabular}
\caption{Comparison with baselines (\%).}
\label{tab:comparison} 
\end{table}

\section{Conclusion}
In this paper, we established a shared task for slot (token)-based in-game toxic language detection, as well as the best-performing model in the shared task, which integrates bi-directional representations, attention residuals, the label forcing technique, and CRF decoders. Experiments indicate that the proposed model is more effective in capturing global information between the semantic components on slot filling than the existing baselines.

\section{Declaration}
This paper constitutes an extended version of the original paper~\cite{11}, primarily incorporating the related work, and providing a detailed elaboration of the methodology of the proposed model.

\bibliography{aaai23}

@article{1,
  title={Using recurrent neural networks for slot filling in spoken language understanding},
  author={Mesnil, Gr{\'e}goire and Dauphin, Yann and Yao, Kaisheng and Bengio, Yoshua and Deng, Li and Hakkani-Tur, Dilek and He, Xiaodong and Heck, Larry and Tur, Gokhan and Yu, Dong and others},
  journal={IEEE/ACM Transactions on Audio, Speech, and Language Processing},
  volume={23},
  pages={530-539},
  year={2014},
  publisher={IEEE}
}

@article{2,
  title={Attention-based recurrent neural network models for joint intent detection and slot filling},
  author={Liu, Bing and Lane, Ian},
  journal={arXiv preprint arXiv:1609.01454},
  year={2016}
}

@inproceedings{3,
  title={Position-aware attention and supervised data improve slot filling},
  author={Zhang, Yuhao and Zhong, Victor and Chen, Danqi and Angeli, Gabor and Manning, Christopher D},
  booktitle={Conference on Empirical Methods in Natural Language Processing},
  year={2017}
}

@inproceedings{4,
  title={Slot Filling with Delexicalized Sentence Generation},
  author={Shin, Youhyun and Yoo, Kang Min and Lee, Sang-goo},
  booktitle={INTERSPEECH},
  pages={2082-2086},
  year={2018}
}

@inproceedings{5,
  title={Improving slot filling in spoken language understanding with joint pointer and attention},
  author={Zhao, Lin and Feng, Zhe},
  booktitle={Proceedings of the 56th Annual Meeting of the Association for Computational Linguistics (Volume 2: Short Papers)},
  pages={426-431},
  year={2018}
}

@inproceedings{6,
  title={CONDA: a CONtextual Dual-Annotated dataset for in-game toxicity understanding and detection},
  author={Weld, Henry and Huang, Guanghao and Lee, Jean and Zhang, Tongshu and Wang, Kunze and Guo, Xinghong and Long, Siqu and Poon, Josiah and Han, Caren},
  booktitle={Findings of the Association for Computational Linguistics: ACL 2021},
  pages={2406-2416},
  year={2021}
}

@inproceedings{7,
  title={Attention-based recurrent neural network models for joint intent detection and slot filling},
  author={Bing, Liu and Ian, Lane},
  booktitle={Interspeech 2016},
  pages={685–689},
  year={2016}
}

@inproceedings{8,
  title={Slot-gated modeling for joint slot filling and intent prediction},
  author={Chih-Wen, Goo and Guang, Gao and Yun-Kai, Hsu and Chih-Li, Huo and Tsung-Chieh, Chen and Keng-Wei, Hsu and Yun- Nung, Chen},
  booktitle={NAACL-HLT 2018},
  pages={753–757},
  year={2018}
}

@inproceedings{9,
  title={A bi-model based rnn semantic frame parsing model for intent detection and slot filling},
  author={Yu, Wang and Yilin, Shen and Hongxia, Jinn},
  booktitle={NAACL-HLT 2018},
  year={2018}
}

@inproceedings{10,
  title={Joint slot filling and intent detection via capsule neural networks},
  author={Chenwei, Zhang and Yaliang, Li and Nan, Du and Wei, Fan and S, Yu, Philip},
  booktitle={Proceedings of the 57th Annual Meeting of the ACL},
  pages={5259–5267},
  year={2019}
}

@inproceedings{11,
  title={In-game toxic language detection: Shared task and attention residuals},
  author={Jia, Yuanzhe and Wu, Weixuan and Cao, Feiqi and Han, Soyeon Caren},
  booktitle={Proceedings of the AAAI Conference on Artificial Intelligence},
  volume={37},
  pages={16238-16239},
  year={2023}
}

\end{document}